\documentclass{article}

\usepackage{microtype}
\usepackage{float}
\usepackage{graphicx}
\usepackage{subfigure}
\usepackage{booktabs} 
\usepackage{amsmath}
\usepackage{amsthm}
\usepackage{amsfonts}
\usepackage{amssymb}
\usepackage{algorithm,algorithmic}
\usepackage{hyperref}

\usepackage[utf8]{inputenc}
\usepackage{amsmath}
\usepackage{amsthm}
\usepackage{amsfonts}
\usepackage{caption}
\usepackage{algorithm}
\usepackage{graphicx}
\usepackage{subfigure}
\usepackage{comment}
\usepackage{multicol}
\usepackage{bbm}
\usepackage[export]{adjustbox}

 \newcommand{\bma}[1]{\mbox{\boldmath $#1$}}

 \newcommand{\bA}{ {\bma{A}} }

 \newcommand{\bH}{ {\mathbf{H}} }

 \newcommand{\bU}{ {\bma{U}} }

 \newcommand{\bGamma}{ {\bma{\Gamma}} }
  \newcommand{\indicator}{\mathbbm{1}}

\usepackage[accepted]{icml2019}

\icmltitlerunning{Parameterized Exploration}

\begin{document}
\setlength{\textfloatsep}{0.1in}
\captionsetup[table]{skip=10pt,labelfont=it,font=small}
\setlength{\abovedisplayskip}{0.1in}
\setlength{\belowdisplayskip}{0.1in}
\twocolumn[
\icmltitle{Parameterized Exploration}



\icmlsetsymbol{equal}{*}

\begin{icmlauthorlist}
\icmlauthor{Jesse Clifton}{ncsu}
\icmlauthor{Lili Wu}{ncsu}
\icmlauthor{Eric Laber}{ncsu}
\end{icmlauthorlist}

\icmlaffiliation{ncsu}{Department of Statistics, North Carolina State University, Raleigh, North Carolina, United States}

\icmlcorrespondingauthor{Jesse Clifton}{jclifto@ncsu.edu}

\icmlkeywords{Machine Learning, ICML}

\vskip 0.3in
]



\printAffiliationsAndNotice{}  

\begin{abstract}
We introduce Parameterized Exploration (PE), a simple family of methods for 
model-based tuning of the exploration schedule in sequential decision problems.  
Unlike common heuristics for exploration, our method accounts for the time horizon 
of the decision problem as well as the agent's current state of knowledge of the 
dynamics of the decision problem.  We show our method as applied to several common 
exploration techniques has superior performance relative to un-tuned counterparts
in Bernoulli and Gaussian multi-armed bandits, contextual bandits, and a Markov decision process based
on a mobile health (mHealth) study.  We also examine the effects of the
accuracy of the estimated dynamics model
on the performance of PE.

\end{abstract}

\section{Introduction}

\par While significant attention has been paid in the reinforcement learning 
literature to domains such as game-playing, in which large amounts of data may be 
generated from a reliable simulator, quantitiative researchers are also interested in 
decision problems in which data are scarce, noisy, and expensive. Examples from the 
statistical literature include the estimation of optimal treatment regimes from 
observational or randomized studies \citep{robins1986,murphyZThree,robins2004optimal,chakraborty2014inference}; just-in-time adaptive 
interventions for mobile health applications \citep{ertefaie2014constructing,luckett2018estimating}; and the control of disease outbreaks 
\citep{meyer2018}.  In such cases, computational efficiency is of secondary 
importance while data-efficiency is paramount.  Moreover,  in such decision problems, 
time horizons may be short and the best-performing exploration strategies must 
account for the number of remaining decisions. 

\par With a view to such problems, we introduce a simple class of algorithms under 
the heading Parameterized Exploration (PE).  Given a class of exploration strategies, 
e.g., $\epsilon$-greedy \citep{watkins,sutton} or upper-confidence bound 
exploration \citep{lai1985asymptotically} 
the PE algorithm simply tunes the exploration schedule using an estimated model of the underlying system dynamics.  This approach leverages the current state of knowledge about the generative model as well as the time horizon.   

\par Perhaps most similar to our algorithm in the recent literature is 
the "Noisy Net," introduced by \cite{fortunato2017noisy}, in which 
exploration in the deep reinforcement learning setting is 
induced by adding parameterized noise to the weights of the neural 
network, and tuning the parameters governing the exploration noise via 
gradient descent.  Our method is similar in spirit to the ``meta-gradient'' 
approach of \citet{xu2018meta} for tuning reinforcement learning hyperparameters.
However, each of these approaches is model-free, while our method leverages
an estimator of the environment dynamics to tune the rate of exploration.
Finally, the literature on Thompson sampling \citep{thompson1933likelihood}
contains several examples of modifying the posterior used to induce 
exploration in order to vary the degree of exploration. 
\citet{chapelle2011empirical}
present empirical results which suggest 
that a modified version of Thompson sampling in which the posterior 
variance is shrunk may outperform standard Thompson sampling, and 
in the context of clinical trials,
\citet{thall2007practical} present a variant of Thompson sampling in 
which randomization probabilities
are adjusted according to the remaining number of patients in the trial.

\section{Setup and notation}
For simplicity, we will consider multi-armed bandits (MABs) in 
presenting the setup and method; the extension of the 
proposed methods to more general sequential decision settings is straightforward.

\par Let $T < \infty$ be the time horizon, $k$ be the number of possible actions, and $D_1, \dots, D_k$ be the unknown reward distributions of each decision with means $\mu_1, \dots, \mu_k$; define $\mu^* = \max_i \mu_i$.  At each time $t = 1, \dots, T$, let $A^t \in \{ 1, \dots, k \}$ denote the selected action and  $U^t \sim D_{A^t}$ the observed reward.  Write the observed sequences of actions and rewards up to time $t$ as $\bA^t = (A^1, \dots, A^t)$ and $\bU^t = (U^1, \dots, U^t)$, and define the history until time $t$ to be $\bH^t = (\bA^t, \bU^t)$.  The goal is to take a sequence of actions such that the expected cumulative regret $\mathbb{E} \Big[ \sum_{t=1}^T \Big( \mu^* - U^t \Big) \Big]$ is minimized.  

\par Define a learning algorithm as a sequence $\bGamma = (\Gamma^1, \dots, \Gamma^T)$, where for each $t$, $\Gamma^t: \mathcal{H}^t \longrightarrow \mathcal{S}_{\mathcal{A}}$, where $\mathcal{S}_{\mathcal{A}}$ is the set of probability distributions over the set $\mathcal{A}$ of actions, and $\mathcal{H}^t$ is the space in which $\bH^t$ takes values.    

\par We consider variants of three of the most popular learning algorithms.  The first is $\epsilon$-greedy: for a decision problem with $k$ actions available at time $t$, this algorithm takes the greedy (estimated-optimal) action with probability $1 - \frac{\epsilon^t}{k}$, and selects actions uniformly at random otherwise.  The second is upper confidence bound (UCB) exploration: this algorithm takes the action which has the greatest upper $(1 - \alpha^t)\times 100\%$ confidence bound on its mean reward.  The third is Thompson sampling:  in classical TS for multi-armed bandits, a sample is taken from the posterior distribution over the mean rewards, and the action
which maximizes these posterior draws is taken \citep{thompson1933likelihood}.  
However, we consider a more general class of TS algorithms. Let 
$C^t_i$ denote a confidence distribution \citep{schweder2002confidence} for the 
mean reward under action $i$ at time $t$; let $\omega_i^t$ denote the mean of 
$C_i^t$.  At each time
$t$ and action $i$ we draw a value $\tilde{\mu}_i^t$ from $C_{i}^t$ and select the action which
maximizes $\omega_{i}^t + \tau^t(\tilde{\mu}_{i}^t-\omega_{i}^t)$ for parameter $\tau^t$;
alternatively, one 
could truncate $C_i^t$ to its $(1-\tau^t)$ highest density region, sample means from this
truncated distribution, and select the action that maximizes the sampled 
means.  Thus, the amount of exploration in the three algorithms is dictated
by the sequences $\{e^t\}_{t=1}^T, \{\alpha^t\}_{t=1}^T$, and 
$\{ \tau^t \}_{t=1}^T$, respectively.

\par We can write each of these learning algorithms in terms of their respective exploration parameters.  For instance, the $\epsilon$-greedy algorithm can be written as 
\begin{align*}
\Gamma_{\epsilon^t}(\bH^t) = \begin{cases}
\arg \underset{i=1, \dots, k}{\max} \bar{U}^t_i, \text{with probability } 1 - \frac{\epsilon^t}{k}, \\
j, \text{with probability } \frac{\epsilon^t}{k}, \text{ for } j=1, \dots, k,
\end{cases}
\end{align*}
where $\bar{U}^t_i$ is the sample mean of rewards observed from arm $i$ until time $t$.  We write the subscript $\epsilon^t$ to emphasize the dependence on the exploration parameter $\epsilon^t$.  More generally, we can write a decision rule with a generic exploration parameter $\eta^t$ as $\Gamma_{\eta^t}$.  

\par In each case, it is clear that the optimal sequence $\{ \eta^t \}_{t=1}^T$ is nonincreasing in $t$, as the value of exploring goes to 0 as we approach the time horizon.  In the next section, we present a simple strategy for tuning the sequence $\{ \eta^t \}_{t=1}^T$ for a given class of learning algorithms.

\section{Parameterized exploration}

\par In order to adaptively tune the rate of exploration, we propose to parameterize the sequence $\{ \eta^t \}_{t=1}^T$ using a family of nonincreasing functions, such that for each $t$, $\eta^t = \eta(T, t, \theta)$ for some $\theta \in \Theta$. We consider the following class of functions:
$\{ \eta(T, t, \theta) =  \frac{\theta_0}{1 + \exp{[-\theta_2(T - t - \theta_1)]}}: \theta=(\theta_0, \theta_1, \theta_2) \in \Theta \}$,
with $\Theta$ chosen such that 
$\eta(T, \cdot, \theta)$ is decreasing for each $\theta \in \Theta$.
Then, if $\Gamma_\eta$ is a decision rule with exploration parameter $\eta$, each value of $\theta$ leads to a learning algorithm $\bGamma_{(\eta, T, \theta)} = (\Gamma_{\eta(T, 1, \theta)}, \dots, \Gamma_{\eta(T, T, \theta)})$.  We will refer to this algorithm as $\bGamma_\theta$, suppressing the dependence on the time horizon $T$ and the class of functions $\eta$.

\par Let $\widehat{\mathcal{M}}^t$ be an estimator of the generative model $\mathcal{M}$ underlying the sequential decision problem at time $t$ --- in the MAB setting, the generative model consists of the reward distributions at each arm, i.e., $\mathcal{M} = (D_1, \dots, D_k)$.  Define $R^T(\theta, \mathcal{M}) = \mathbb{E}_{\mathcal{M}, \theta} \Big[ \sum_{t=1}^T \Big( \mu^* -  U^t \Big) \Big]$ to be the cumulative regret until the horizon $T$ if actions are chosen according to $\bGamma_\theta$ and the true generative model is $\mathcal{M}$.  Then, at each time step $t$ we can solve $\hat{\theta}^t = \arg\min_{\theta \in \Theta} R^T(\theta, \widehat{\mathcal{M}}^t)$ 
and take the action $A^t = \Gamma_{\eta(T, t, \hat{\theta}^t)}(\bH^t)$.

\par However, in early episodes, point estimates $\widehat{\mathcal{M}}^t$ of $\mathcal{M}$ may be 
of low quality due to insufficient training data; for instance, if 
the variances of the reward distributions are simultaneously under-estimated and the 
ordering of estimated reward means is incorrect, 
this method may lead to 
under-exploration and therefore long sequences of suboptimal actions.  (We observed such behavior in preliminary 
simulation experiments.)
In order to account for 
uncertainty in $\widehat{\mathcal{M}}^t$, we can instead minimize the expected value of the above 
objective 
quantity against a confidence distribution $C^t$ for $\mathcal{M}$ at time $t$.  That is, we can 
solve
$\hat{\theta}^t = \arg\min_{\theta \in \Theta} \mathbb{E}_{\widetilde{\mathcal{M}} \sim C^t} R^T(\theta, \widetilde{\mathcal{M}})$.
%
%
to get $\hat{\theta}^t$ and associated decision rule at each time.  This variant of PE is 
presented in Algorithm \ref{alg:pe}.

\begin{algorithm}[h!]
\caption{Parameterized exploration for MABs}
\label{alg:pe}
\begin{algorithmic}
\STATE \textbf{Input} Function class $\{ \eta(T, \cdot, \theta) : \theta \in \Theta \}$; decision rule class $\{ \Gamma_\eta : \eta \geq 0 \}$; reward distributions $\{D_i \}_{i=1}^k$; time horizon $T$
\STATE $\bA^0 = \{1, \dots, k\}$
\STATE $\bU^0 = \{U^0_1 \sim D_1, \dots, U^0_k \sim D_k\}$
\STATE $\bH^0 = \{ \bA^0, \bU^0 \}$
\FOR{t = 0, \dots, $T-1$}
\STATE Obtain $C^t$ from $\bH^t$
\STATE $\hat{\theta}^t \gets \arg\underset{\theta \in \Theta}{\max} \: \mathbb{E}_{\widetilde{\mathcal{M}} \sim C^t} R^T(\theta, \widetilde{\mathcal{M}})$
\STATE $A^t \sim \Gamma_{\eta(T, t, \hat{\theta}^t)}(\bH^t)$
\STATE $U^t \sim D_{A^t}$
\STATE $\bH^{t + 1} \gets \bH^t \cup \{ A^t, U^t \}$
\ENDFOR
\end{algorithmic}
\end{algorithm}

\section{Simulations}
We present comparisons of tuned and un-tuned variants of the learning
algorithms discussed above in multi-armed bandits (MABs), a normal-linear contextual
bandit,
and a continuous-state
Markov decision process (MDP).  In the MDP case, 
we modify the objective to maximize
cumulative reward rather than minimize cumulative regret.
In each case,
we carry out the requisite optimizations using Gaussian process
optimization, implemented in the Python package 
\texttt{BayesianOptimization} 
\citep{BayesOpt}.

\subsection{Bernoulli multi-armed bandit}
We consider Bernoulli MABs with 2, 5,and 10 arms.
Table \ref{table:regrets1} compares mean cumulative regrets of 
tuned and untuned variants of $\epsilon$-greedy, UCB, and 
Thompson sampling.  We also report the performance of finite 
horizon Gittins index \citep{kaufmann2018bayesian}
 in the table; the finite horizon Gittins index 
 policy is an important baseline given that
 it is approximately Bayes-optimal and therefore represents
 another approach to incorporating the full state of 
 knowledge about the environment as well as the time
 horizon into the exploration.
 For each family of learning algorithms,
 the tuned variant is competitive with or outperforms
 the un-tuned variants. The tuned algorithms are also
 competitive with the Gittins index policy in most cases.

\subsection{Gaussian multi-armed bandit}
Following \citet{kuleshov2014algorithms}, we test our algorithms 
on Gaussian MABs with reward means in $[0, 1]$ and different 
choices of variances.  We set the time horizon at $T=50$ given
our focus on short-horizon decision problems. Table \ref{table:regrets}
displays the cumulative regrets for tuned and un-tuned variants of
$\epsilon$-greedy, UCB, and Thompson sampling, 
with 2, 5, and 10 arms.
The tuned variants outperform un-tuned counterparts
across many settings, and when they do not have the lowest sample
mean regret are within approximately one
standard error of 
the methods which do perform
best in sample mean. 

\begin{table}[h!]
    \centering
    \setlength\tabcolsep{3.75pt}
        \caption{Bernoulli MAB: Mean cumulative regrets and standard errors (in parentheses) of different methods for different number of arms over 192 replicates. $t=1,2,\dots,50.$}
    \label{table:regrets1}
    \begin{tabular}{c|c|c|c}
  \hline
  \hline
    &\multicolumn{3}{c}{Mean Cumulative Regrets (SE)}\\
\cline{2-4}
 Methods & 2 arms & 5 arms & 10 arms \\ 
  \hline
  \hline
  Tuned $\epsilon$-greedy
    & \textcolor{blue}{1.64(0.08)} & \textcolor{blue}{3.08(0.25)} & \textcolor{blue}{4.60(0.35)}	\\ 
   $\epsilon$-greedy ($\epsilon=0.05$) & 2.78(0.32) & \textcolor{red}{2.67(0.20)} & 4.58(0.29) \\ 
    $\epsilon$-greedy ($\epsilon=0.1$) &2.54(0.24) &2.76(0.19)&4.96(0.27) \\ 
    $\epsilon_t=0.5/t$ & 2.40(0.30) & 2.87(0.22) & \textcolor{red}{4.43(0.29)}\\ 
   \hline 
   Tuned TS
   & \textcolor{blue}{1.02(0.15)} & \textcolor{blue}{2.88(0.20)} & \textcolor{blue}{4.33(0.27)}\\ 
 TS & 2.39(0.20) & 3.41(0.18) & 6.45(0.25) \\ 
 $\eta^t=1/t$ & 2.14(0.30) & \textcolor{red}{2.57(0.21)} & 4.50(0.33)\\
 
   \hline
    Tuned UCB
    & \textcolor{blue}{1.08(0.13)} & \textcolor{blue}{3.11(0.21)} & \textcolor{blue}{4.21(0.28)}\\ 
 UCB ($\alpha=0.05$) & 1.89(0.15)  &  3.28(0.18) & 5.92(0.25)\\ 
 $\alpha^t=0.5-0.45/t$ & 2.44(0.34)&3.15(0.24)&4.62(0.29) \\
 
 \hline
 Gittins index & 1.50(0.20) & 2.72(0.18) & 4.37(0.24)\\
   \hline

\end{tabular}

\end{table}

\begin{table*}[ht]
    \centering
    \caption{Gaussian MAB: Mean cumulative regrets and standard errors (in parentheses) of different methods for different $\sigma$ generative models over 192 replicates ($\sigma=0.1$), 384 replicates ($\sigma=1$). $t=1,2,\cdots,50.$  Peformance of tuned methods in blue.  Performance
    of method with best mean performance in red where different
    from the tuned variant.}
    \begin{tabular}{c|c|c|c|c|c|c}
  \hline
  \hline
    &\multicolumn{6}{c}{Mean cumulative regrets (SE)}\\
\cline{2-7}
&\multicolumn{2}{c}{2 arms}&\multicolumn{2}{|c|}{5 
arms}&\multicolumn{2}{c}{10 arms}\\
\cline{2-7}
 Methods & $\sigma=1$ & $\sigma=0.1$& $\sigma=1$ & $\sigma=0.1$& $\sigma=1$ & $\sigma=0.1$ \\ 
  \hline
  \hline
  Tuned $\epsilon$-greedy
    &\textcolor{blue}{4.76(0.30)} & 	\textcolor{blue}{0.14(0.08)} & \textcolor{blue}{5.92(0.31)} & 
    \textcolor{blue}{0.89(0.11)} & \textcolor{blue}{8.43(0.27)} & \textcolor{blue}{0.74(0.06)}\\ 
   $\epsilon$-greedy ($\epsilon^t=0.05$) & \textcolor{red}{4.62(0.30)} & 0.39(0.04) & 5.64(0.26) & 0.83(0.09) & 8.65(0.27) & 1.28(0.06)\\ 
    $\epsilon$-greedy ($\epsilon^t=0.1$) & 4.78(0.28) & 0.74(0.04) & \textcolor{red}{5.37(0.24)} & 0.98(0.09) & 8.80(0.25) & 1.98(0.07)\\ 
    $\epsilon$-greedy ($\epsilon^t=0.5/t$) &  4.98(0.31) & 0.32(0.02) & 5.46(0.27) & \textcolor{red}{0.77(0.10)} & \textcolor{red}{8.35(0.28)} & 1.07(0.07)\\ 
   \hline 
   Tuned TS
   & \textcolor{blue}{4.76(0.32)} & \textcolor{blue}{0.06(0.01)} & \textcolor{blue}{6.00(0.29)} & 
   \textcolor{blue}{0.72(0.09)} & \textcolor{blue}{8.66(0.27)} & \textcolor{blue}{0.71(0.05)} \\ 
 TS $(\tau^t = 1)$ & 5.09(0.29) & 0.25(0.01) & 7.14(0.23) & 1.28(0.04) & 10.30(0.21) & 2.18(0.05)\\ 
 TS ($\tau^t=1/t$) & 5.29(0.35) & \textcolor{red}{0.05(0.01)} & 6.16(0.32) &0.90(0.11) & 8.78(0.29) & 0.74(0.05)\\
 
   \hline
    Tuned UCB
    & \textcolor{blue}{5.41(0.35)} &  \textcolor{blue}{0.02(0.01)} & \textcolor{blue}{5.52(0.31)} & \textcolor{blue}{0.44(0.09)} & 
    \textcolor{blue}{8.33(0.28)} & \textcolor{blue}{0.44(0.06)}\\ 
 UCB ($\alpha^t =0.05$) & 5.37(0.34) &  0.08(0.01) & 6.80(0.32) & 0.70(0.04) & 9.52(0.25) & 1.27(0.04)\\ 
 UCB ($\alpha^t=0.5-0.45/t$) & \textcolor{red}{5.02(0.34)} & 0.09(0.08) & 5.86(0.29) & 0.85(0.11) & 8.61(0.29) & 0.64(0.05)\\
 
   \hline
\end{tabular}
    
    \label{table:regrets}
\end{table*}

\subsection{Contextual bandits}
In the contextual bandit setting, we considered the two-arm normal-linear contextual 
bandit of \citet{lei2017actor}, based on the Heartsteps applications for increasing
physical activity \citet{klasnja2015microrandomized}. In order to tune the 
exploration parameter, we fit a correctly specified multivariate normal model
for the context distribution.
The results are displayed in Table
\ref{tab:cb}.
\begin{table}[ht!]
\caption{Contextual bandit performance of $\epsilon$-greedy variants; 96 replicates of $T=50$ time steps.}
    \centering
    \begin{tabular}{c|c}
        Methods & Mean cumulative regret (SE)  \\ \hline
        Tuned $\epsilon$-greedy  & \textcolor{blue}{1.31(0.07)} \\
        $\epsilon = 0.05$        & 1.47(0.10) \\
        $\epsilon_t = t^{-1}$    & 2.12(0.08) \\
        $\epsilon_t = 0.5t^{-1}$ & 1.95(0.08) \\
        $\epsilon_t = 0.8^t$     & 2.13(0.08)\\
        \hline
    \end{tabular}
    
    \label{tab:cb}
\end{table}
\subsection{Controlling glucose: a continuous-state MDP}
In this experiment, we consider an MDP in which
the states are continuous. We simulate cohorts of patients with type 1 
diabetes using a generative model based on the mobile 
health study of 
\citep{maahs2012outpatient}.
We only consider the action of whether to use insulin, so the 
action space is $\mathcal{A}=\{0,1\}$. The covariates observed for 
patient $i$ at time $t$ is average blood glucose level, total 
dietary intake, and total counts of physical activity, denoted 
by $(Gl_i^t,Di_i^t,Ex_i^t)$ respectively.  The
action taken at time $t$ is denoted as $A^t$.
Defining states 
$S^t = (Gl^t,Di^t,Ex^t,Gl^{t-1},Di^{t-1},Ex^{t-1},A^{t-1})^\intercal $,
glucose levels evolve according to the second-order autoregressive
(AR(2)) process
    $Gl^t = \beta S^{t-1} + e^t$,
where $e^t \sim N(0,5^2)$ and $\beta=(10, 0.9, 0.1, -0.01, 0.0, 0.1, -0.01, -10, -4)$; 
and $Di^t\sim N(0, 10^2)$ with 
probability $0.6$, otherwise $Di^t=0$; similarly,  
$Ex^t\sim N(0, 10^2)$ with probability 
$0.6$, otherwise $Ex^t=0$.
Thus the dynamics are Markovian with states 
$S^t$. 
The reward at each time step is given by 
$U^t \equiv \indicator( Gl^t < 70)[ -0.005(Gl^t)^2 + 0.95 
Gl^t - 45] + \indicator(Gl^t \geq 70)[ -0.0002(Gl^t)^2
+ 0.022 Gl^t - 0.5
]$, which decreases as $Gl^t$ departs from normal glucose 
levels.
(This is a continuous variant of the discrete reward
function used in \citet{luckett2018estimating}.)

\par We simulate data for $n=15$ 
patients and time horizons of 25 and 50.  
In each case we use an $\epsilon$-greedy learning algorithm,
where the greedy action 
is given by the argmax of the estimated
conditional expected reward function, fit using random
forest regression.
We examined three approaches to estimating the glucose
transition dynamics.  First, we fit a correctly-specified
AR(2) linear model using ordinary
least squares.
Second, we fit an 
(incorrectly-specified) AR(1) linear model in which
glucose depends only the glucose, food intake, activity
level, and action at the previous time.  Finally,  
we model the conditional probability distribution of glucose at time $t$
given the glucose, food, activity, and treatments from times $t-1, t-2$ using a two-step procedure
similar to that described in \cite{hansen2004nonparametric}: First, we estimate the 
conditional mean of glucose using a random forest estimator,
and then the full conditional distribution of the residuals of the fitted models
by the ratio of the kernel density estimators of the joint distribution of glucose and covariates
to the marginal density of the covariates; the relevant bandwidths were selected using
cross-validation.  In each of these 
transition model estimators, 
activity and food are
assumed to be i.i.d. over time points and patients,
and their distributions are estimated by their
empircal distributions.

\par The mean cumulative regrets incurred by each
$\epsilon$-greedy variant for $T=25,50$ are displayed
in Table \ref{table:np-glucose}.  While the linear models,
both correct and misspecified, do considerably better
than each of the other methods, tuning with the 
nonparametric conditional density estimator performs
worst or among the worst at both time horizons.

\begin{table}[h]
    \centering
    \caption{Comparison of $\epsilon$-greedy variants in the 
    glucose problem in terms of mean cumulative reward (MCRew).
    96 replicates for T=25,
    192 replicates for T=50 (as these
    were considerably higher-variance).
    Conditional glucose distribution estimated using
    i) (correctly specified) AR(2) linear model, 
    ii) (incorrectly specified) AR(1) linear model,
    and iii) (correctly specified) AR(2) nonparametric conditional
    density estimator.}
    \begin{tabular}{c|c|c}
    Time horizon & Method & MCRew (SE) \\
    \hline
      Tuned (AR(2) linear) 
        & \textcolor{blue}{-13.08(0.55)} 	\\ 
      & Tuned (AR(1) linear) 
        & -13.39(0.84) \\
      & Tuned (AR(2) NP)
        & -30.40(1.97)    	\\ 
      T=25& $\epsilon^t=0.05$ & -17.57(0.53)   \\ 
      & $\epsilon^t = t^{-1}$ & -20.02(1.25) \\
      & $\epsilon^t = 0.5t^{-1}$ & -17.80(0.90)\\
      & $\epsilon^t = 0.8^t$ & -22.93(1.79) \\
       \hline
    Tuned (AR(2) linear)
        & \textcolor{blue}{-16.92(0.42)}  	\\ &
    Tuned (AR(1) linear)
        & -25.10(3.45) \\
    & Tuned (AR(2) NP)
        & -52.80(4.45)   	\\ 
     T=50 & $\epsilon^t = 0.05$ & -39.56(2.55)  \\ 
      & $\epsilon^t = t^{-1}$ & -57.23(4.24)  \\
      & $\epsilon^t = 0.5t^{-1}$ & -33.83(3.41)  \\
      & $\epsilon^t = 0.8^t$ & -52.93(4.45)  \\
       \hline

    \end{tabular}
    
    \label{table:np-glucose}
\end{table}

\section{Future work}
While we have only considered model-based tuning of exploration parameters,
the method presented here could be used to tune other 
hyperparameters such as the
discount factor used in the estimation of the optimal policy, or 
modify the objective function to improve (for instance) the 
operating characteristics of statistical tests of comparisons between
patients receiving different treatments (see 
\citet{williamson2017bayesian} for an alternative apprach to trading off
expected in-sample rewards and statistical power in the context of 
clinical trials).
Another direction for
future work is constructing diagnostics to identify if the estimated 
system dynamics are of sufficiently high-quality to justify 
their use in tuning, as evidenced by the poor 
performance of the nonparametric conditional density
estimator, which suffers from high variance.
Finally, a theoretical regret analysis of our 
method is called for.

\bibliographystyle{plainnat}
\bibliography{aiStats_copy}

\end{document}